\def\paperTitle{QPoser: Quantized Explicit Pose Prior Modeling for Controllable Pose Generation}

\def\PriorName{QPoser}
\def\VQVAEName{MS-VQVAE}
\def\VQVAENameLong{Multi-head Sequential Vector Quantized Variational Autoencoder}
\def\AEName{GLIF-AE}
\def\AENameLong{Global-Local Intertwined Features Auto Encoder}

\newif\ifreview 
\newif\ifarxiv 
\newif\ifcamera 
\newif\ifrebuttal 

\PassOptionsToPackage{svgnames,dvipsnames}{xcolor}
\documentclass[journal]{vgtc}  

\onlineid{0}

\vgtccategory{Research}

\vgtcpapertype{please specify}

\title{\paperTitle{}}

\author{%
  Yumeng Li,
  Yaoxiang Ding,
  Zhong Ren\thanks{Corresponding author.}, and 
  Kun Zhou
}

\authorfooter{
  \item
  	Y. Li, Y. Ding, Z. Ren and K. Zhou are with the State Key Lab of CAD\&CG, Zhejiang University
}

\abstract{%
  Explicit pose prior models compress human poses into latent representations for using in pose-related downstream tasks. A desirable explicit pose prior model should satisfy three desirable abilities: 1) correctness, i.e. ensuring to generate physically possible poses; 2) expressiveness, i.e. ensuring to preserve details in generation; 3) controllability, meaning that generation from reference poses and explicit instructions should be convenient. Existing explicit pose prior models fail to achieve all of three properties, in special controllability. To break this situation, we propose QPoser, a highly controllable explicit pose prior model which guarantees correctness and expressiveness. In QPoser, a multi-head vector quantized autoencoder (MS-VQVAE) is proposed for obtaining expressive and distributed pose representations. Furthermore, a global-local feature integration mechanism  (GLIF-AE) is utilized to disentangle the latent representation and integrate full-body information into local-joint features. Experimental results show that QPoser significantly outperforms state-of-the-art approaches in representing expressive and correct poses, meanwhile is easily to be used for detailed conditional generation from reference poses and prompting instructions.
}


\teaser{
    \centering
\includegraphics[width=1.0\textwidth]{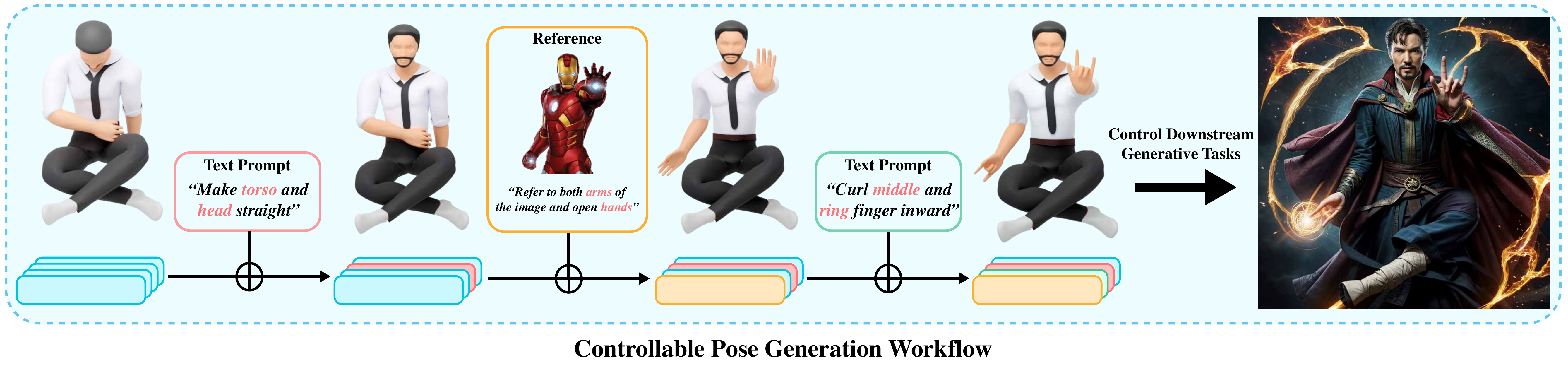}
\captionof{figure}{Demonstration of the controllable pose generation abilities.
Our pose prior model facilitates nuanced control from body-level pose to intricate finger-level adjustments using both prompt and reference inputs. These refinements are further guided by embodied information, ensuring natural and accurate output. As content generations are often human centered, controllable pose generation can further bring controllability to downstream generative tasks, such as image generation.}
\label{fig:teaser}
}

\renewcommand{\manuscriptnotetxt}{}

\graphicspath{{figs/}{figures/}{pictures/}{images/}{./}} 

\usepackage{graphicx}	
\usepackage{amsmath}	
\usepackage{amssymb}	
\usepackage{booktabs}
\usepackage{times}
\usepackage{microtype}
\usepackage{epsfig}
\usepackage[table,xcdraw,dvipsnames]{xcolor}
\usepackage{caption}
\usepackage{float}
\usepackage{placeins}
\usepackage{color, colortbl}
\usepackage{stfloats}
\usepackage{enumitem}
\usepackage{tabularx}
\usepackage{xstring}
\usepackage{multirow}
\usepackage{xspace}
\usepackage{url}
\usepackage{subcaption}
\usepackage[numbers,sort&compress]{natbib}
\usepackage[hang,flushmargin]{footmisc}

\begin{document}


\firstsection{Introduction}
\label{sec:intro}

\maketitle

Human pose modeling is an essential preliminary step for varies human-related visual generative tasks~\cite{controlnet, KIM200845_Virtual_Try_On, Wang_2023_CVPR_Transfer_Pose, huang2023dreamwaltz, hong2022avatarclip,azadi2023textconditionalposepersonal,Zhang_2020_CVPR_Generate_People_In_Scene,zhang2022wanderings}. However, even though human pose can be modeled explicitly by geometry in the $SO^3$ joint space~\cite{KodekShoulderElbow,HATZE1997128JointMobility}, this primitive representation is sparse and non-smooth, such that most samples in the joint space are not plausible human poses, and traversing the joint space can lead to implausible in-between poses~\cite{SMPL-X:2019,tiwari22posendf,advposeprior}. This could lead to significant negative effect on the downstream generation quality.

Pose priors~\cite{SMPL,SMPL-X:2019,advposeprior,hmrKanazawa17,kocabas2019vibe,ci2022gfpose,tiwari22posendf}, which model the prior distribution of human poses, are proposed to address these issues. Compared with naive geometry representation, a proper pose prior model can provide valuable additional information of the pose space by assigning more plausible poses with higher probabilities. Among them, {\it explicit pose priors}~\cite{SMPL-X:2019,advposeprior} further require to build bijective mappings from poses to explicit manifold spaces, which are usually vector embeddings. The major advantages of explicit pose priors are two-fold: on one hand, the generality of vector embedding representation makes them the plug-in choice to model poses in various downstream tasks~\cite{Wang_2023_CVPR_Transfer_Pose}.

\begin{figure*}[tp]
    \centering
    \includegraphics[width=\linewidth]{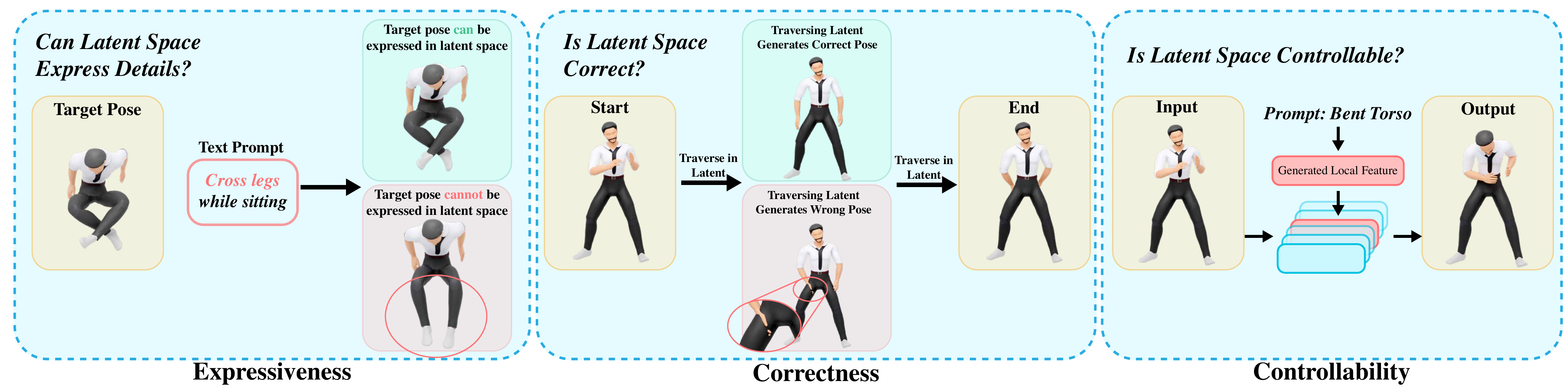}
    \caption{\textbf{Abilities of Explicit Pose Priors:}
    We demonstrate the abilities mentioned in Section \ref{sec:intro} with examples.
    In application, all these abilities need to be obtained simultaneously, because it is unacceptable if: 1) target pose cannot be expressed in latent space, thus it cannot be generated. 2) output poses are incorrect. 3) detailed control, such as reference images or textual descriptions, cannot be performed.}
    \label{fig:abilities}
\end{figure*}

On the other hand, their explicit representation space can serve as a solution space for generative tasks. This is particularly crucial for recently advanced AI-based content generation tasks~\cite{Wang_2023_CVPR_Transfer_Pose, huang2023dreamwaltz, hong2022avatarclip,azadi2023textconditionalposepersonal,Zhang_2020_CVPR_Generate_People_In_Scene,zhang2022wanderings,Aliakbarian_2022_CVPR_Sparse_Observation}, as the primitive joint space is too sparse to serve as a generation solution space. However, to achieve a higher level of generation ability, we argue that a desirable explicit pose prior model should possess the following abilities as illustrated in Figure \ref{fig:abilities}:

1. \emph{Expressiveness}: meaning the model should ensure perfect generation when the target pose is a plausible one. As the latent representation serves as the target solution space, the target pose has to be accurately expressed in the latent space in order to be generated.

2. \emph{Correctness}: meaning that the model should ensure to omit impossible poses. Furthermore, correctness also indicates the ability to keep smoothness in sequential generation, which can be evaluated with qualitative measurements such as pose interpolation.  

3. \emph{Controllablity}: meaning that the model should support reference-based and prompting-based conditional generation ability, as shown in Figure \ref{fig:teaser}. This ability not only requires explicit representation of poses, but also requires the representations to be {\it disentangled} and {\it embodied} so as to allow local changes w.r.t. full-body configurations.

The motivation of our work is based on the following issue of the existing explicit pose prior models:

{\it There is not existing explicit pose prior model achieving all above abilities, in special controllability.}

To address this challenge, we propose \PriorName{}, an explicit latent pose prior model targeting at achieving the best of all worlds.
This is achieved through two innovative architectures: 1) \VQVAENameLong{} (\VQVAEName{}), a vector-quantized variational autoencoder (VQ-VAE) model~\cite{vqvae} for representing the pose space in a highly {\it expressive} and {\it stable} way. We propose a multi-head architecture to map the full pose into distributed codes. This allows to extract rich features from human poses in an effective manner. 2) \AENameLong{} (\AEName{}), which disentangles the latent representation and integrates full-body information into local-joint features. Such {\it embodied} feature representation not only ensures controllability w.r.t. the demands on local changes, but also {\it correctness} due to the injection of full-body information.

Our experiments show that compared with existing baselines, QPoser significantly improves the generation expressiveness and correctness. More importantly, QPoser achieves desirable controllability by allowing detailed generation from reference and prompting instructions.

\section{Related Work}
\label{sec:related}
In this section, we discuss two categories of mainstream pose prior models: explicit and implicit ones, as well as existing controllable pose prior models.

\emph{Implicit pose priors.}
This category of pose priors implicitly model the pose distribution with whole neural network model parameters, without explicit latent representations of human poses. Some previous works often train these prior models for one specific tasks, such as human pose and mesh recovery~\cite{hmrKanazawa17, kocabas2019vibe} and 3D pose hypotheses generation~\cite{li2019generatinghypotheses}, while recent works utilize architectures like the score network~\cite{ci2022gfpose} and neural distance field~\cite{tiwari22posendf} as a uniformed method to learn pose distributions across a few downstream tasks like motion denoising, mesh recovery and pose hypotheses generation. These implicit models perform well in their designated tasks, but their inherent architectures restrict their use in downstream generative tasks that require plug-in representation of poses. More importantly, achieving controllability is complicated and challenging for them due to their implicit representations. 

\emph{Explicit pose priors.}
State-of-the-art explicit pose priors predominantly utilize generative neural models to obtain the vector-based embedding spaces of poses. The seminal work of VPoser~\cite{SMPL-X:2019} incorporates variational autoencoder (VAE)~\cite{kingma2013variationalautoencoder} for pose modeling, improving classical method which is based on the Gaussian mixture model (GMM)~\cite{SMPL}. VPoser has been widely used as explicit pose prior for various downstream generative tasks~\cite{Wang_2023_CVPR_Transfer_Pose,azadi2023textconditionalposepersonal,huang2023dreamwaltz,hong2022avatarclip,Zhang_2020_CVPR_Generate_People_In_Scene,Aliakbarian_2022_CVPR_Sparse_Observation}. Due to the convenience of adopting vector representations, explicit prior models can usually be used interchangeably in downstream tasks, which is one of their significant benefits. Recently, the adversarial parametric pose prior~\cite{advposeprior} is also proposed based on the generative adversarial network (GAN) architecture~\cite{GAN}. Despite the appealing properties of these existing models, based on thorough experiments, we discover that even though correctness is usually guaranteed, their expressiveness is not that ideal, meanwhile more importantly, the controllability is still lacking.

{\it Controllable pose priors.} It is worth noting that motion priors have also been investigated~\cite{motionvae,petrovich21actor,rempe2021humor,Ao2023GestureDiffuCLIP,tangSIG23RSMT}, highlighting the pivotal role of prior models in human pose related tasks. These motion priors also emphasize the significance of controllability in generation. For example, \citeauthor{zhang2022motiondiffuse} endeavored to blend different body parts for realistic composition, and \citeauthor{weiyu23GenMM} applied a generative matching method to complete motion when given lower-body motion, achieving a certain level of controllability. However, the targets of these motion priors are addressing specific motion generation tasks. This is different from our purpose for designing controllable pose prior models for general downstream tasks, which is still beyond the ability for existing approaches.

\section{Method}
\label{sec:method}

Our objective is to create an explicit pose prior model with a representation space characterized by expressiveness, correctness, controllability, as well as a distributed vectorized language-like structure for seamless integration with the transformer architecture~\cite{vaswani2017attention}, leveraging its expressiveness and the potential for unified integration with diverse control signals. This is particularly relevant as an increasing number of modalities are adopting transformer architectures for multi-modal integration~\cite{zhu2023minigpt, chen2023minigptv2,jiang2023motiongpt}.

Many recent works create such latent representation utilizing vector-quantized variational autoencoder (VQ-VAE)~\cite{vqvae} ranging from image~\cite{vqdiffusion}, audio~\cite{jukebox}, to motion generation ~\cite{Ao2023GestureDiffuCLIP,tevet2022motionclip,zhang2022motiondiffuse}. It operates through encoding the input, quantizing the encoded features using codebook, then decoding it back. While it may seem straightforward to consider VQ-VAE as a drop-in replacement for variational autoencoders (VAE) used by VPoser, the implementation is actually non-trivial since VQ-VAE usually requires the input data to be splittable and redundant to be compressed, which is not the case for pose modeling, such that the primitive inputs are low-dimensional vectors in the $SO^3$ space. The key observation is that, even though this space seems to be low-dimensional, both plausible and implausible poses can exist within it. Therefore, we still need to do information compression on it to obtain a manifold space which includes only plausible poses. To the best of our knowledge, how VQ-VAE can be applied on such situation is not studied before in the literature.
We conclude the major challenges into three key points: 1) Can we effectively apply VQ-VAE to data representations that are neither splittable nor dense, such as human pose? 2) Can we disentangle the latent representation to enable control over different body parts? 3) Can we ensure that the disentangled local features maintain correctness by respecting the global embodied information?

\begin{figure*}[tp]
    \centering
    \includegraphics[width=\linewidth]{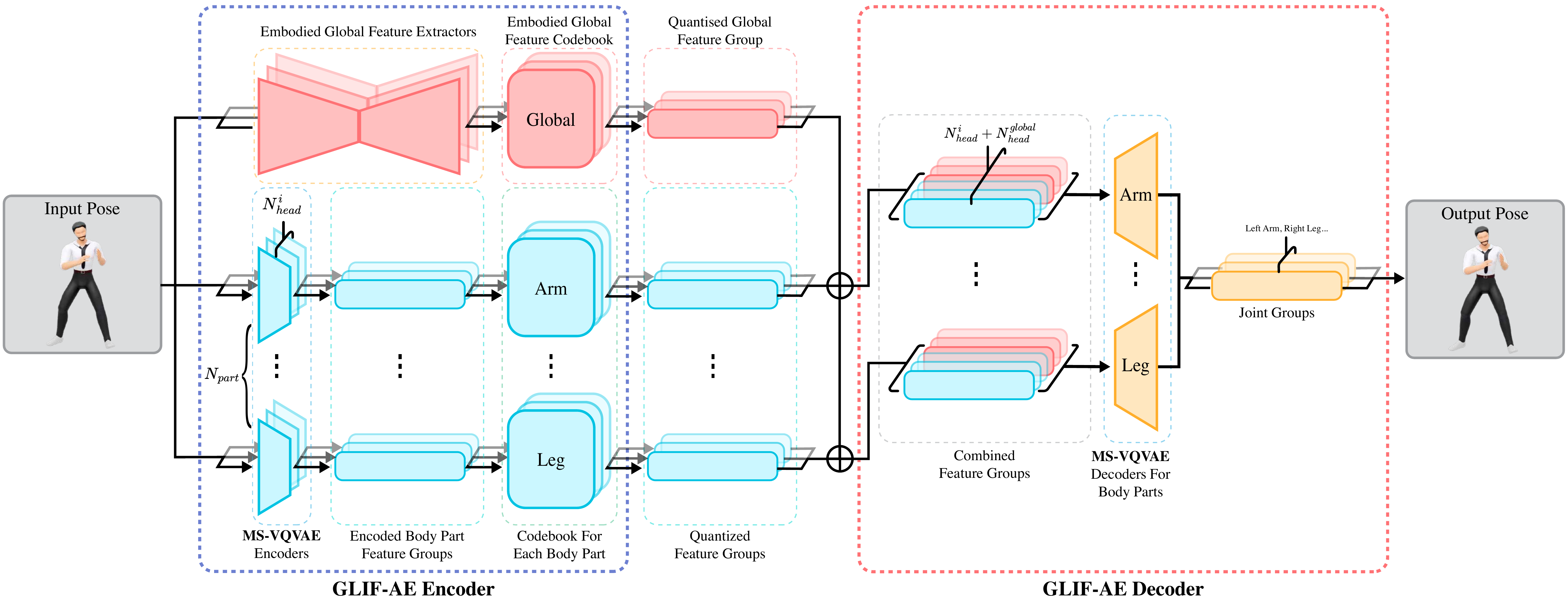}
    \caption{\textbf{\PriorName{} Architecture:}
    Full pose is processed into partial feature groups with specialized encoder heads. Simultaneously, global embodied info extraction via 'Embodied Global Feature Extractors' and subsequent quantization using codebooks, followed by feature group concatenation and decoding for the synthesis of joint rotations and final output generation.}
    \label{fig:architecture}
\end{figure*}

To address all these challenges, we introduce \PriorName{}, a quantized and disentangled explicit pose prior model. The central components of QPoser is a multi-head vector quantized autoencoder and a global-local feature integration mechanism, as illustrated in Figure \ref{fig:architecture}. When presented with an input pose, \PriorName{} encodes it into distributed and vectorized latent representations. These latent representations are organized with explicit semantic meanings, permitting meaningful partial adjustments while adhering to the constraints imposed by the embodied information. Once a desired latent representation is obtained, it can be decoded back into the joint space pose for further utilization. Below we dive into details of the two central components.

\subsection{Multi-Head Vector-Quantized Encoding}
\label{sec:VQVAEArch}

Given our objective of constructing a language-like latent feature using an auto-encoder architecture, we must address the challenge of applying VQ-VAE to low-dimensional dense data representations, such as human poses. We design \VQVAENameLong{} (\VQVAEName{}) to handle this challenge.

Unlike traditional VQ-VAE which involves a single encoder, \VQVAEName{} employs a series of unique encoders. Each of these encoder heads is responsible for compressing the full pose, which has dimensions of $N_{joint} \times d_{joint}$, into a compact representation of dimension $d_{code}$. 
By stacking these unique encoders, a single dense and non-splittable human pose can be encoded into a distributed vectorized latent feature. More specifically, given the pose $P$, we train our model with encoder-quantization-decoder structure:
\begin{equation*}
    Z = \mathcal{E}(P),\quad Z_q = \mathcal Q(Z), \quad  \hat P = \mathcal{D}(Z_{q}),
\end{equation*}
where $\mathcal{E}, \mathcal{D}, \mathcal Q$ are encoder, decoder, and vector-quantized modules respectively. Instead of trying to find a method to split the data representation, we use the full pose $P$ as the input of a multi-layer perceptron (MLP) network $F$, and compress it into one single latent code $z_{code}=F(P)$.

We stack this structure for $N_{head}$ times, each with a different MLP network which is considered as one head, producing a final latent representation $Z$ such that
\begin{equation*}
    Z = \left[z_{code}^{i}\right]_{i=1}^{N_{head}} = \left[F^{i}(P)\right]_{i=1}^{N_{head}}.
\end{equation*}
By adopting this multi-head architecture, we extract a series of distributed features from the pose. We then quantize the code with code book $C$:
\begin{equation*}
    Z_{q} = \left[z_{q}^{i}\right]_{i=1}^{N_{head}} = \left[C(z_{code}^{i})\right]_{i=1}^{N_{head}}.
\end{equation*}
Now we have a quantized and distributed latent representation $Z_{q} \in \mathbb{R}^{N_{head} \times d_{code}}$, extracted from pose data $x \in \mathbb{R}^{N_{joints} \times d_{joint}}$, as the input to the decoder $\mathcal{D}$.

\subsection{Global-Local Embodied Representation}
\label{sec:AEArch}

While \VQVAEName{} has succeeded in achieving an expressive and stable latent representation, controllability is still not achieved. To introduce controllability, we must disentangle the latent representation for different body parts within the latent space while maintaining the relationships between each body part and the global embodied information to ensure correctness. This is achieved by the \AENameLong{} (\AEName{}) architecture.

We assume that each body part is governed by a distribution denoted as $P(z_{part} | z_{embodied})$. Here, $z_{part}$ signifies the latent representation for a specific body part, and $z_{embodied}$ represents the global embodied information. Given a body part $P_{part}$ and the complete pose $P$, our goal is to autonomously discover the latent representation $z_{part}$ for the body part, as well as the latent representation of the embodied information $z_{embodied}$.

To achieve this, we leverage the entire pose as the input, rather than splitting it. This allows us to provide ample information for the model to learn. We implement the disentanglement process by designing a unique decoder structure. Within this structure, we divide the encoded latent representation into distinct groups, with one special group representing the global embodied information, and each of the other groups representing a different body part. Subsequently, we assign each group of latent representations to be decoded solely for their corresponding body part, together with the embodied global information.

This innovative architecture enables us to disentangle the latent representations effectively, thereby promoting controllability while preserving correctness in our generative model. More specifically, we train a group of body part encoders
\begin{equation*}
    \left[z_{part}^{i}\right]_{i}^{N_{part}} = \left[\mathcal{E}_{part}^{i}(P)\right]_{i}^{N_{part}}
\end{equation*}
as well as an embodied global feature encoder
\begin{equation*}
    z_{embodied} = \mathcal{E}_{embodied}(P).
\end{equation*}
All the body part encoders and the embodied global feature encoder accept the same full body pose $P$ and have no major difference in structure. Their main difference is how their produced latent features are being used by the decoders:

\begin{equation*}
    \left[P_{part}^{i}\right]_{i}^{N_{part}} = \left[\mathcal{D}_{part}^{i}(z_{part}^{i}, z_{embodied})\right]_{i}^{N_{part}},
\end{equation*}
where the final output pose $P$ can be represented as $P = \sum_{i=1}^{N_{part}}P_{part}^{i}$. By designing the architecture in this way, we encourage the model to only encode the global embodied info into $z_{embodied}$ and only encode the embodied info irrelevant body part info into $z_{part}^i$. Decoders are then responsible for understanding the conditional distribution $P(z_{part} | z_{embodied})$ and produce the output pose $P$. Experiments show that this architecture can model the complex relationship between body part and the global embodied info, ensuring the correctness of a body part within the pose. This is especially obvious in detailed partial latent manipulations.

\subsection{QPoser}
\label{sec:QPoser}

Combining the above two architectures, we can now introduce the whole \PriorName{} model. As shown in Figure \ref{fig:architecture}, we train an encoder-decoder pair following the \AEName{} architecture, while using \VQVAEName{} for all the sub auto-encoders. We break body pose into 6 parts: Head, Torso, Left Arm, Right Arm, Left Leg, Right Leg. Symmetric parts share the same codebook, including both arms and legs. Based on \AEName{}'s requirement, we add a global embodied feature encoder and its corresponding codebook as well. Finally, we attach the quantized global embodied features to each body part's quantized features, then decode them to the joint rotations for their corresponding body parts, and rearrange these joints to construct the final output pose. There are $N_{part}$ sub encoders in total, with $\left[N_{head}^{i}\right]_{i}^{N_{part}}$ encoder heads. We also break the hand representation into 5 fingers and train a separate hand pose prior using the same method.

During training, we apply two losses, i.e. reconstruction loss and commitment loss, similar to standard VQ-VAE:
\begin{equation*}
\mathcal{L} = \mathcal{L}_{recon} + \lambda \mathcal{L}_{commit},
\end{equation*}
where
$   \mathcal{L}_{recon} = \| P - \hat{P}\|^2_2, \mathcal{L}_{commit} = \| z - z_q\|^2_2,$
and $\lambda$ the weighting hyper-parameter set as $1$ in the experiments. The overall model is then trained with standard stochastic gradient descent optimizer for optimization.

\subsection{Workflow for Solving Downstream Tasks}
\label{sec:downstream}

We design an experimental iterative workflow for controllable generation to showcase the capabilities and potential of our prior model, as depicted in Figure \ref{fig:teaser}. At the heart of this workflow lies our pose prior's explicit and disentangled latent representation. The following modules are integrated with the prior to generate and manipulate the latent representations, enabling correct and controllable generation:

\textbf{Unconditional Generation}: We explore the potential of our language-like latent representation by employing a decoder-only Transformer architecture to estimate the masked latent code. Qualitative experiments show that this approach can generate natural and diverse human poses, underscoring the expressiveness and correctness of our pose priors. It also demonstrates the potential for our latent representation to be treated as a foreign language and aligned with other modalities.

\textbf{Control by Reference}: Referencing is a common task in human workflows. With our latent representation, we can accurately incorporate local features from references into the target pose. This integration seamlessly combines with the current embodied information and preserves the desired features. Additionally, human images can serve as references, using state-of-the-art human 3D pose regression methods like \cite{cai2023smplerx}.

\textbf{Control by Prompting}: We also conduct experiments on text-conditioned generation by mapping text features created by SBERT \cite{reimers2019sentencebert} into our latent space using a multi-head encoding architecture. The results indicate that text features can effectively generate corresponding latent pose features (see Figure \ref{fig:generation}). While this process still requires some prompt engineering and may not handle all text inputs equally well due to current dataset limitations, it showcases the significant potential of our latent representation to integrate seamlessly with other modalities.

To facilitate practical use, we have developed an intuitive system that leverages these modules for generating and manipulating latent features in an iterative manner. Users can provide further instructions to modify poses within the loop until they are satisfied. These generated poses can then be employed in downstream tasks, such as controlling digital human avatars or generating images.

\begin{figure*}[tp]
    \centering
    \includegraphics[width=0.85\linewidth]{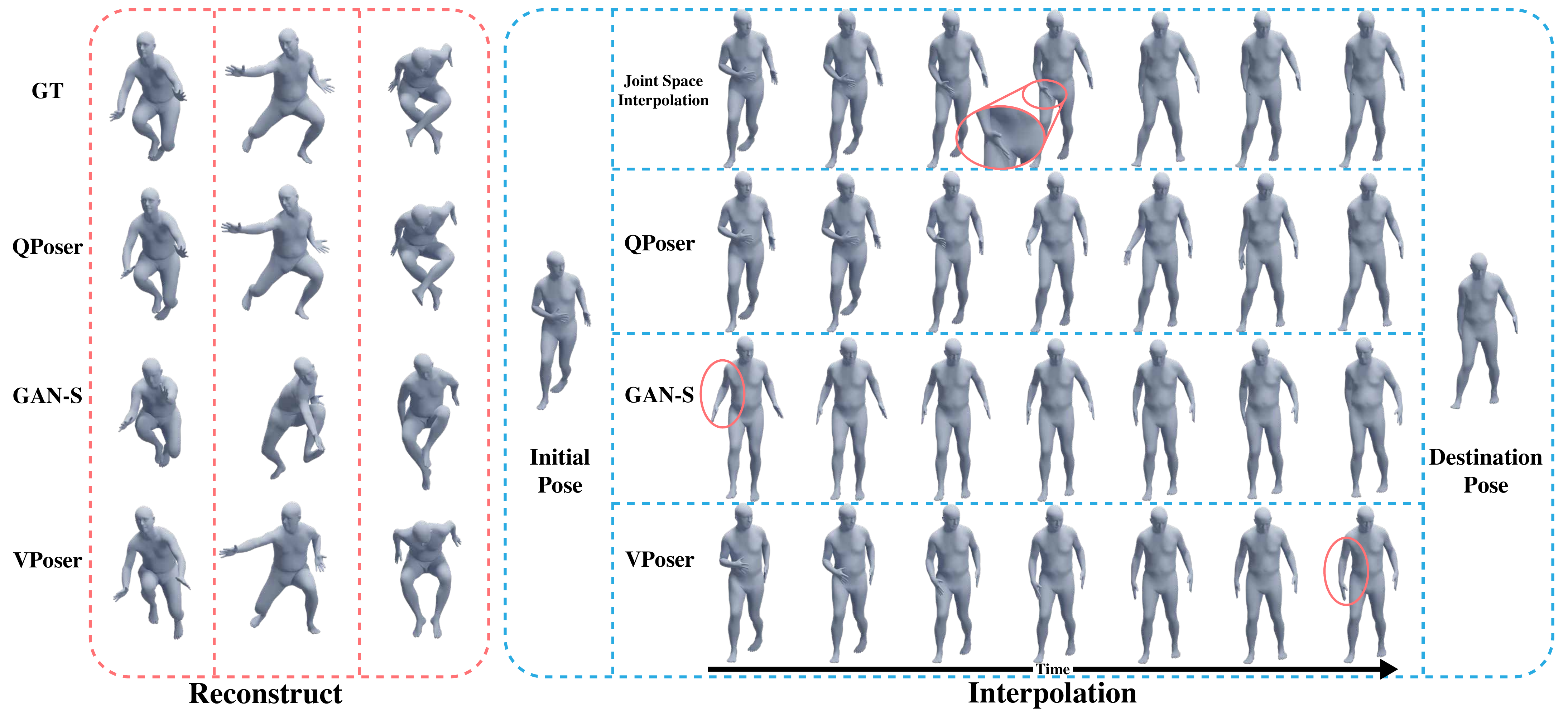}
    \caption{\textbf{Comparison:}
    We show the qualitative comparison for different pose priors in expressiveness and latent correctness. Our models can accurately express input poses' details in it's latent, demonstrated by it's reconstruction accuracy. We also show that our latent produce correct, smooth and natural in-between interpolations, even when compared with GAN based models. Note that this example also highlighted the importance of pose priors, as joint space interpolation leads to obvious wrong results.
    }
    \label{fig:comparison}
\end{figure*}

\section{Experiments}
\label{sec:experiments}
In this section, we compare the performance of QPoser with state-of-the-art explicit pose priors, VAE-based VPoser~\cite{SMPL-X:2019} and GAN-based adversiral parametric pose prior~\cite{advposeprior} under the following benchmark tasks:
\begin{itemize}
    \item 
    {\it Verifying expressiveness.} We evaluate the expressiveness ability under the {\it latent reconstruction} task, where the pose prior models are required to perfectly reconstruct plausible poses with high precision. Furthermore, the expressiveness can also be evaluated through whether poses with more diversity can be generated under {\it random sampling}. 
    \item 
    {\it Verifying correctness.} We evaluate the generation correctness under the {\it sequential interpolation} and {\it local modification} tasks, which requires the model to generate smooth sequence of poses or conduct proper local changes. The correctness can also be evaluated under {\it random sampling} to test whether the model can indeed avoid implausible generation results.
    \item 
    {\it Verifying controllability.} We evaluate the newly introduced controllability of \PriorName{} qualitatively with {\it local modification} and {\it controlled generation} tasks.
\end{itemize}

\subsection{Implementation Details}
We collected a dataset comprising 20 million poses from the AMASS motion dataset~\cite{AMASS:ICCV:2019} for training, validation and testing. We separately trained a hand prior model with the same architecture as our pose prior, using the InterHand2.6M dataset~\cite{Moon_2020_ECCV_InterHand2.6M} with MANO~\cite{MANO:SIGGRAPHASIA:2017} hand representation annotated by NeuralAnnot~\cite{Moon_2022_CVPRW_NeuralAnnot,Moon_2023_CVPRW_3Dpseudpgts}. Both pose data and hand data use quaternion joint rotation as representation.

\begin{table}[t]
\centering

\begin{tabular}{lcccc}
\toprule
Reconstruct Error MPJAE(°) & Mean$\downarrow$ & Std$\downarrow$\\
\midrule
VPoser & 7.47 & 5.57\\
GAN-S & 10.62 & 8.51\\
\PriorName{}(ours) & \textbf{2.62} & \textbf{1.64}\\
\midrule
Escalated Error MPJAE(°) & Mean$\downarrow$ & Std$\downarrow$\\
\midrule
VPoser & 13.06 & 12.94\\
GAN-S & 22.64 & 21.81\\
\PriorName{}(ours) & \textbf{0.34} & \textbf{1.49}\\
\bottomrule
\end{tabular}

\caption{
\textbf{Pose Reconstruction:} We evaluate the expressiveness of pose priors' latent representation using reconstruction error. If a pose prior is expressive enough, it should accurately preserve the details of the original pose. In addition, a pose prior should also accumulate less error as human workflows are iterative. This can be evaluated by recursively reconstructing a pose and check it's escalated error. We show that our prior significantly outperforms current models. Especially, our prior has a close to zero accumulated error, which makes it practical to be applied in generation workflows. Note that expressiveness matters only if the prior model learns a meaningful latent representation since auto-encoders can "cheat" by not compressing the input. Our further experiments shows that our latent representation is correct and smooth, proving that our model does compress the input into a meaningful latent.
}
\vspace{-0.05in}
\label{tab:recon_error_and_stab}
\end{table}

\begin{figure}[tp]
    \centering
    \includegraphics[width=\linewidth]{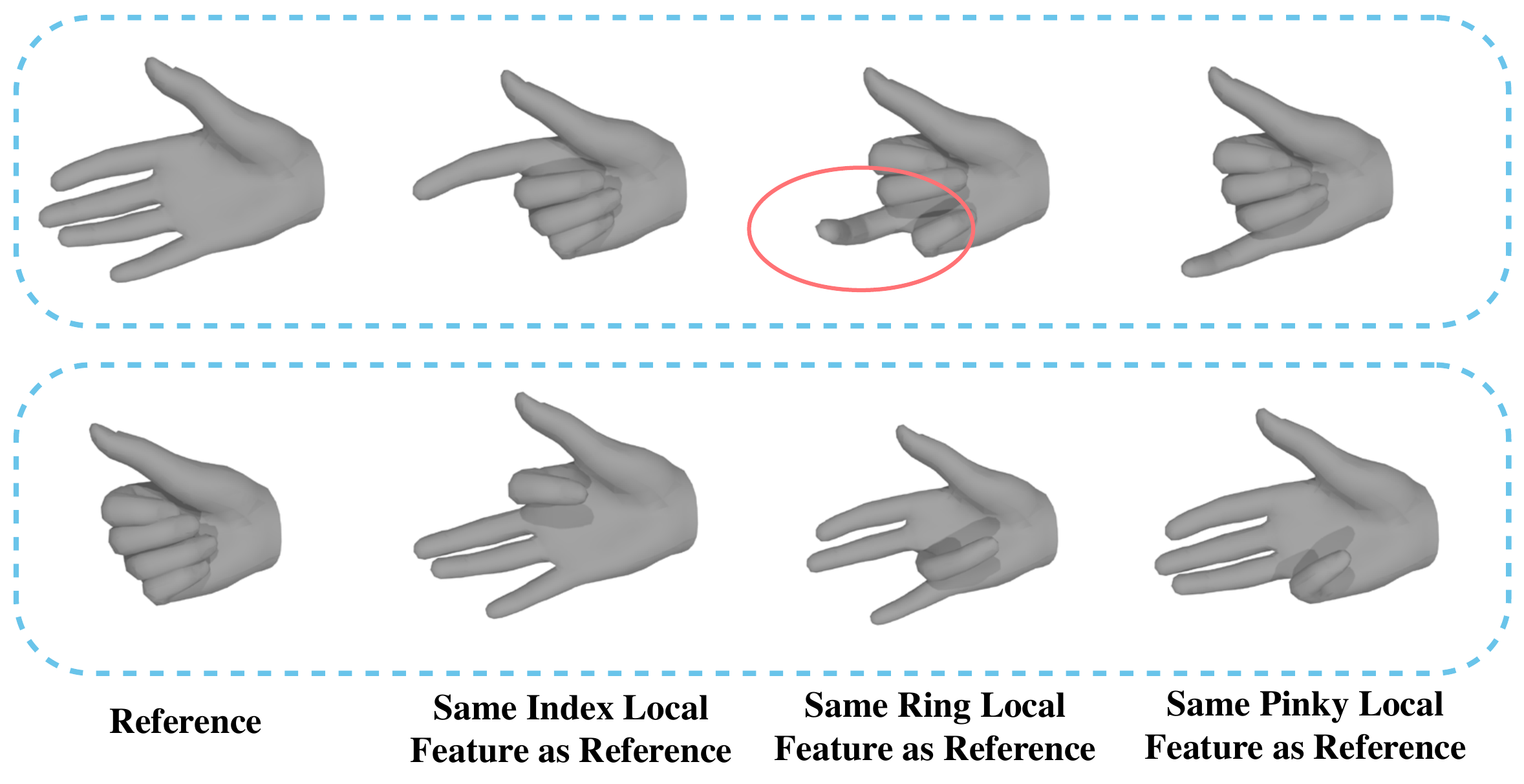}
    \caption{\textbf{Partial Modification:}
    We demonstrate how identical local feature will provide different outcome under different embodied info. This effect is especially noticeable on complex body components, such as hand. Fingers can only curl or straighten to a certain degree that it's possible under the current hand gesture, for example, ring finger cannot fully straighten when forming a fist, and when the other fingers are straightened, the curled finger will not make contact with the palm.}
    \label{fig:partial}
\end{figure}

\subsection{Latent Reconstruction}

To assess expressiveness of QPoser, we reconstruct the original pose from latent representation and measure the error introduced in this process. The evaluation metrics include Reconstruct Error and Escalated Error, both quantified in terms of Mean Per Joint Position Error (MPJAE).

\textbf{Reconstruct Error}: An expressive latent should preserve most details. Our prior model exhibits significantly lower errors when mapping the latent representation back to the joint space, highlighting its enhanced expressiveness. Detailed quantitative results can be found in Table \ref{tab:recon_error_and_stab}, and qualitative comparisons are illustrated in Figure \ref{fig:comparison}.

\textbf{Escalated Error}: Iterative mapping between joint and latent space is required in human workflow, as user needs to view the current result in joint space, and provide further instructions for adjustments to be made in latent space. This highlights the necessity of preventing error amplification during iterations. We define escalated error as the difference between error after iterations ($error_{iteration}$) and error after reconstruction ($error_{reconstruct}$) for a fair comparison. Our pose prior demonstrates substantially lower escalated error after 50 iterations, as demonstrated in Table \ref{tab:recon_error_and_stab}. Our further experiment showed that the escalated error remained stable and negligible (0.33°) even after 1000 iterations, affirming the practicality and safety of our prior model for iterative applications.

\subsection{Sequential Interpolation}

We conducted an interpolation test to assess the correctness and smoothness of our prior model's latent space. The qualitative results, as depicted in Figure \ref{fig:comparison}, demonstrate that \PriorName{} excels in creating smooth and accurate interpolations between poses, outperforming other pose priors.

\subsection{Random Sampling}

\begin{figure}
    \centering
    \includegraphics[width=\linewidth]{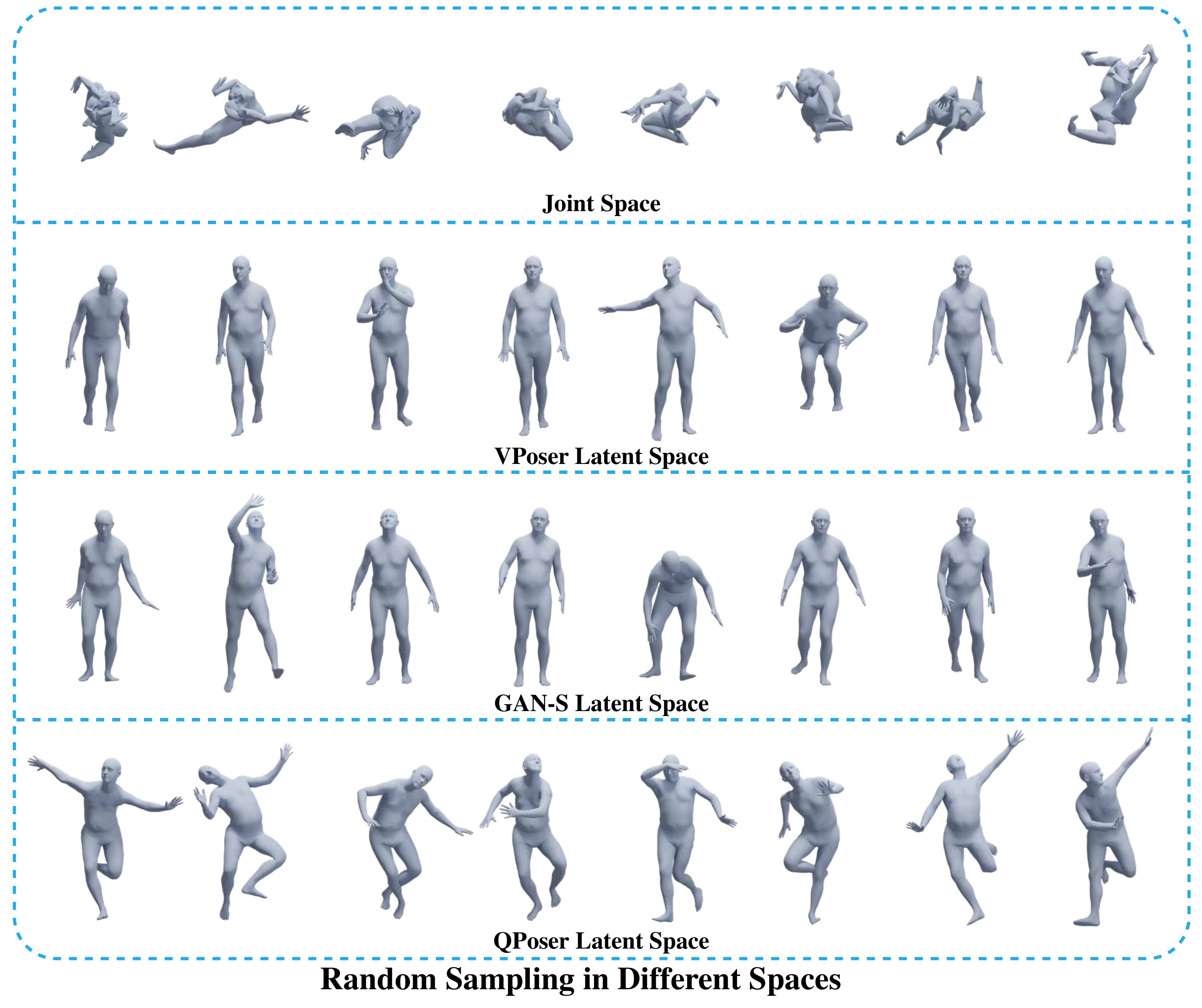}
    \caption{\textbf{Space Comparison:}
    We conducted random sampling across various spaces. Notably, sampling in joint space leads to obvious wrong results. While sampling in latent spaces for explicit pose priors all lead to correct poses, we observe that sampling in QPoser's latent space yielded more diverse poses, underscoring its superior expressiveness. When generating diverse results is not desirable, our pose priors can also perform unconditioned generation based on designated distribution, as discussed in Section \ref{sec:downstream} and demonstrated in Figure \ref{fig:generation}.
    }
    \label{fig:sampling_comparison}
\end{figure}

We evaluate correctness and expressiveness by randomly sampling from different representation spaces. A correct and expressive space should produce correct and diverse output when being sampled randomly. As shown in Figure \ref{fig:sampling_comparison}, while maintaining correctness, our prior produces diverse poses, highlighting its expressiveness. Additionally, our prior model does not exhibit a tendency to generate a 'mean' pose nor is it constrained by the distribution of the training set. This resolves challenges encountered by VAE and GAN-based approaches~\cite{advposeprior,tiwari22posendf}.

\begin{figure*}[tp]
    \centering
    \includegraphics[width=0.75\linewidth]{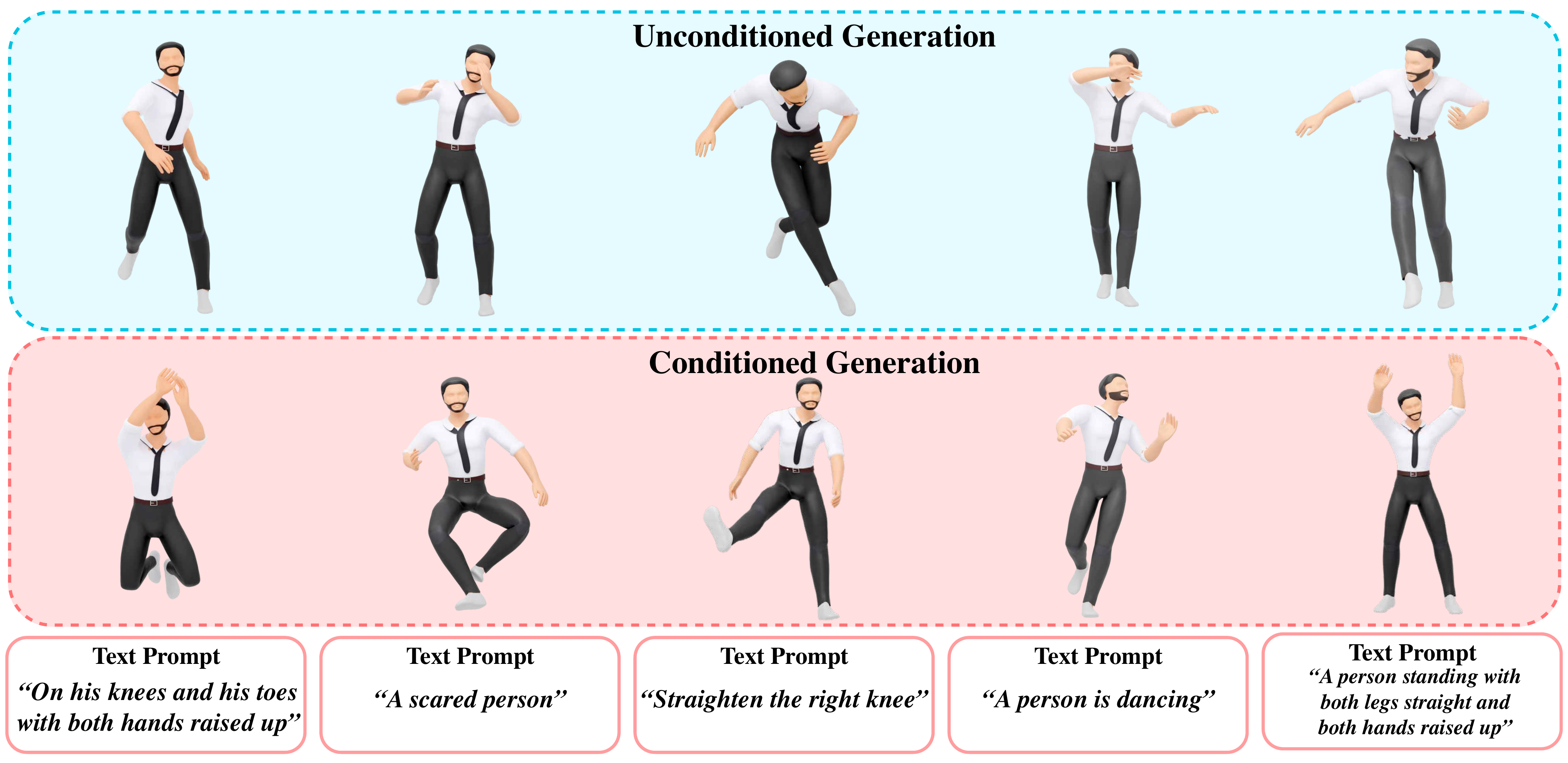}
    \caption{\textbf{Generation:}
    We demonstrate the unconditioned and conditioned generation ability of our pose prior as discussed in Section \ref{sec:experiments}.
    }
    \label{fig:generation}
\end{figure*}

\subsection{Local Modification}

We further demonstrate our prior model's disentanglement for body parts' latent representation, as well as its correctness by performing qualitative tests on latent modification. As shown in Figure \ref{fig:partial}, we show that our model can correctly model the conditional distribution of local parts under the constraint of embodied global info: when given the local feature to fully open the finger, under the embodied global info of a fist, ring finger cannot be fully opened, while index finger and pinky finger can easily straighten. This ability is essential for ensuring that correctness is still preserved when introducing controllability by separating the latent representation.

\subsection{Controlled Generation}

The sequential and language-like latent representation of our \PriorName{} enables seamless integration with other architectures, including transformer~\cite{attention_is_all_you_need}. To demonstrate the correctness and feature richness of our latent representation, we constructed a simple decoder-only architecture for unconditioned generation. This showcases the potential to integrate QPoser with transformer-based models.

Furthermore, we illustrate the straightforward integration of various modalities by directly mapping text features into our latent space for conditioned generation by using text-pose pairs from PoseScript dataset~\cite{posescript}. Qualitative results are presented in Figure \ref{fig:generation}, with additional results available in the supplementary materials.

\subsection{Ablation Study}
\label{subsec:ablation}
In this section, we further conduct ablation studies to verify the effectiveness of MS-VQVAE and GLIF-AE modules.
\begin{table}[t]
\centering
\scalebox{0.95}{
\begin{tabular}{lcccc}
\toprule
Reconstruct Error MPJAE(°) & Mean$\downarrow$ & Std$\downarrow$\\
\midrule
1 Head & 14.96 & 11.62\\
8 Heads & 9.70 & 6.53\\
8 Heads with \AEName{} & 9.09 & 6.03\\
32 Heads & 5.15 & 3.23\\
32 Heads with \AEName{}& 3.82 & 2.46\\
63 Heads & 5.06 & 3.19\\
63 Heads with \AEName{}(final) & \textbf{2.62} & \textbf{1.64}\\

\midrule
Escalated Error MPJAE(°) & Mean$\downarrow$ & Std$\downarrow$\\
\midrule
1 Head & 0.62 & 2.76\\
8 Heads & 0.52 & 2.17\\
8 Heads with \AEName{} & 0.94 & 3.17\\
32 Heads & 0.69 & 3.20\\
32 Heads with \AEName{}& 0.50 & 1.64\\
63 Heads & 0.91 & 4.39 \\
63 Heads with \AEName{}(final) & \textbf{0.34} & \textbf{1.49}\\
\bottomrule
\end{tabular}
}

\caption{
\textbf{Ablation for \VQVAEName{} and \AEName{}:} Our ablation tests show that multi-head vector quantized autoencoder is the key to the expressiveness and stability. With only one head, posterior collapse occurred. Note that we cannot apply \AEName{} on a single head. As the number of encoder head grows, the expressiveness of the model increases accordingly. The low escalated error is mainly contributed by our quantized latent, as it remains low even when expressiveness is bad. We also find that \AEName{} can further boost the expressiveness and stability of \VQVAEName{} in most cases.
}
\vspace{-0.05in}
\label{tab:ablation_multi_head}
\end{table}
\begin{figure}
    \centering
    \includegraphics[width=1\linewidth]{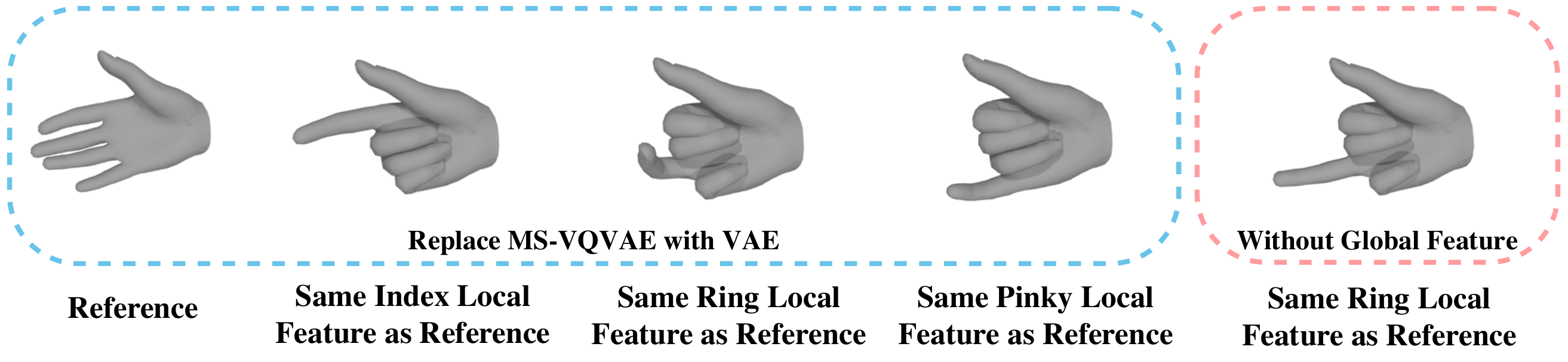}
    \caption{\textbf{Ablation for \AEName{}:}
    We conducted an ablation test for our proposed \AEName{} architecture. Our findings demonstrate the versatility of this architecture, as it can be applied not only to our proposed \VQVAEName{} but also to the VAE architecture, which yields similar qualitative results. Additionally, we examined the ablation results when the global feature was not integrated into the architecture. Notably, when the local feature is left unbounded, it results in entirely implausible outcomes for not respecting the embodied info.}
    \label{fig:ablation_global_local}
\end{figure}

\begin{enumerate}
\item
{\it \VQVAEName{}.} We performed ablation tests to evaluate the effectiveness of \VQVAEName{}, as presented in Table \ref{tab:ablation_multi_head}. The results affirm that our proposed multi-head architecture successfully mitigates the posterior collapse problem and serves as a crucial element in achieving an expressive latent representation. 

\item
{\it \AEName{}.} Ablation tests were conducted to assess the significance of the global-local design in \AEName{}. As illustrated in Figure \ref{fig:ablation_global_local}, our findings underscore that the global-local architecture plays a pivotal role in ensuring that the local features respect the embodied information. When this design is not applied, the model predominantly focuses on the local feature, generating implausible outcomes that disregard the embodied information. We also find that \AEName{} can boost the expressiveness and stability of \VQVAEName{} in most cases, as shown in Table \ref{tab:ablation_multi_head}.

\end{enumerate}

\section{Limitations and Future Work}
\label{sec:future}

Currently, there is a challenge in aligning text and pose due to limitations in available datasets and methods, impeding the effectiveness of text-conditioned generation. In recent researches regarding modal alignment with language, such as vision~\cite{zhu2023minigpt, chen2023minigptv2} and motion~\cite{jiang2023motiongpt}, all modalities are transformed into language-like representations, and then unified with Transformer. Since our pose prior introduces a novel approach to transform human pose into such language-like representation, this innovation opens up the potential for seamless integration of pose with other modalities. It offers the opportunity to enhance feature extraction for various downstream tasks and improve the alignment between human pose and language, which has wide-ranging applications and can be further explored.

\section{Conclusion}
\label{sec:conclusion}

In this paper, we introduced \PriorName{}, an expressive pose prior featuring a language-like vector quantized latent representation. It incorporates an explicitly disentangled latent representation to enable detailed control, while ensuring their compatibility with the global embodied information. Our experiments demonstrate that our model achieves significantly improved expressiveness compared to existing state-of-the-art pose priors and enables a higher level of controllability. We also showcase its easy integration with Transformer architecture and other feature space for generative tasks, revealing the feature-rich nature of our latent representation which empowers controllability. We also developed a prototype workflow for text-conditioned pose generation and modification, emphasizing the superiority and practicality of our pose prior model, revealing the appealing potential of QPoser for further exploration in pose control with multi-modal integration and alignment with human language.

\bibliographystyle{ieeenat_fullname}
\bibliography{qposer}

\clearpage
\appendix 

\section{Implementation Details}

\subsection{Data}

We conducted sampling on a subset of 20 million poses derived from the AMASS motion dataset~\cite{AMASS:ICCV:2019}. This dataset was partitioned into training, validation, and test sets with an 80\%, 10\%, and 10\% distribution, respectively.

To handle hand data, we utilized annotated MANO~\cite{MANO:SIGGRAPHASIA:2017} data from the InterHand2.6M dataset~\cite{Moon_2020_ECCV_InterHand2.6M}, representing hand annotations in the MANO~\cite{MANO:SIGGRAPHASIA:2017} format. This data was segregated into training, validation, and test subsets following the same 80\%, 10\%, and 10\% distribution.

Our approach involved leveraging SMPL-H representations from the AMASS dataset; however, we only considered 21 joints, indexed from 1 to 21. The reason for this selection was due to the majority of the dataset containing static hand representations within SMPL-H. Recognizing that hand pose is relatively independent of human body pose, we pursued a distinct training regimen for the hand pose prior. This hand pose prior was then incorporated into our overarching pose prior, resulting in a comprehensive human pose prior model that encompasses detailed hand features.

Our model was trained employing a normalized quaternion joint representation, ensuring positivity to circumvent ambiguity inherent in quaternion representations. This methodology aimed to enhance the robustness and clarity of our model's learned representations.

\subsection{Architecture}

Our model uses a latent representation divided into codes for different body regions: 3 for the head, 12 each for the torso, arms, and legs, and 3 for global information. These codes correspond to unique MLP encoders.

There are a total of 5 codebooks—one each for the head and global information (with 8 codes), and for the torso, arms, and legs (each with 32 codes). Notably, both arms and legs share the same codebook. The dimensionality of each code is set at 16 to maintain consistency across all body parts and global data.

\begin{figure}[tp]
    \centering
    \includegraphics[width=\linewidth]{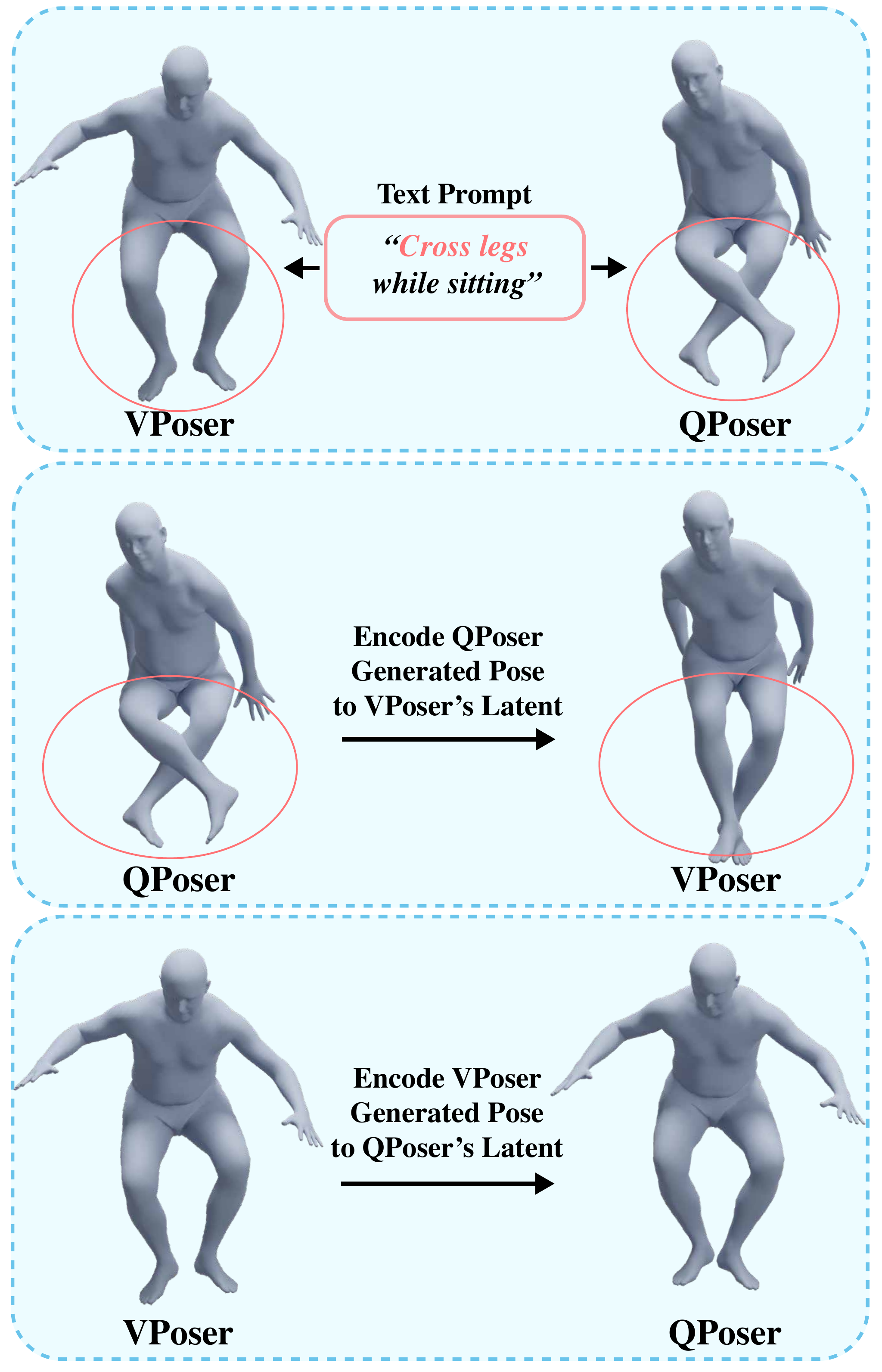}
    \caption{\textbf{Expressiveness Matters:} 
    The expressiveness of a latent space in explicit pose priors plays a crucial role in generative tasks as it serves as the solution space. Demonstratively, conditioned generation encounters limitations when the sought outcome doesn't reside within the solution space provided by VPoser. In contrast, \PriorName{} exhibits significantly higher expressiveness, enabling it to handle such scenarios adeptly. To substantiate that this limitation is due to latent expressiveness rather than the conditional generation process, we encoded pose generated by \PriorName{} into VPoser's latent space and observed that VPoser indeed struggles to represent such pose. This reaffirms the influence of latent expressiveness on the model's capability to represent diverse and complex poses within the solution space. In contrast, pose generated by VPoser can be encoded into QPoser's latent space accurately.
    }
    \label{fig:expressiveness_matters}
\end{figure}

\subsection{Baseline}

We employed the pre-trained VPoser~\cite{SMPL-X:2019} model and assessed both its original and fine-tuned versions using our dataset split. The version exhibiting the best performance was selected as the baseline. It's evident that expanding the expressiveness of VPoser while preserving compression capabilities is unattainable due to its inherent auto-encoder design and the latent dimension being relatively large compared to its input size.

For the GAN-based pose prior~\cite{advposeprior}, as the pre-trained model was unavailable, we trained it using provided code and configuration on our data split. Additionally, a Predictor was trained to map poses back to the GAN's latent space for comparison with our other pose priors. We employ the spherical sampling strategy, identified as the most effective method in the original paper.

\section{Experiment Details}

\subsection{Interpolation}

We encode poses into each pose prior's latent space and perform linear interpolation between the starting and destination pose latents. The interpolated latents are directly fed to the decoder in our model \PriorName{}, bypassing the codebook for continuous results. Supplementary video demonstrations display some interpolation examples, showcasing the effectiveness of our approach.

\subsection{Latent Sampling}
\label{sec:supp_latent_sampling}

For VPoser, random latent samples are drawn from a normal distribution. Meanwhile, for the GAN-based model, we implement the spherical sampling strategy. As for our specific pose prior, we randomly select codes from the respective codebooks to generate poses.

\subsection{Workflow}

Our experimental workflow leverages the controllability offered by our pose prior's latent space. To achieve prompt-based editing, we employ a masking generator to determine modifications required for specific body parts based on the given prompt. When provided with a text prompt, a text encoder extracts distinct features for different body parts from the given pose. For reference images, the main human pose is extracted using image-based human pose regression. Subsequently, our pose prior's encoder encodes this pose into distinct features for different body parts. These extracted features, along with the original poses' features, are combined according to the generated mask. Then, utilizing our decoder, which has learned the relationship between these features, the joint space pose corresponding to these latent features is generated. This comprehensive workflow enables the manipulation of poses based on various prompts and references, effectively utilizing the capabilities of our pose prior's latent space for controllable pose editing. We showcase a downstream application of the generated pose by employing ControlNet \cite{controlnet} in its 'Hard Edge' mode. The 3D poses we generate offer rich 3D information to the image generator, ensuring correct and controllable image generation.\footnote{The image reference of the dancer is created by two-dreamers from Pexels, and the reference of IronMan is created by Marvel. All images are generated using Realistic Vision V5.1 by SG\_161222.}

\section{More Results}

\begin{figure*}
    \centering
    \includegraphics[width=\linewidth]{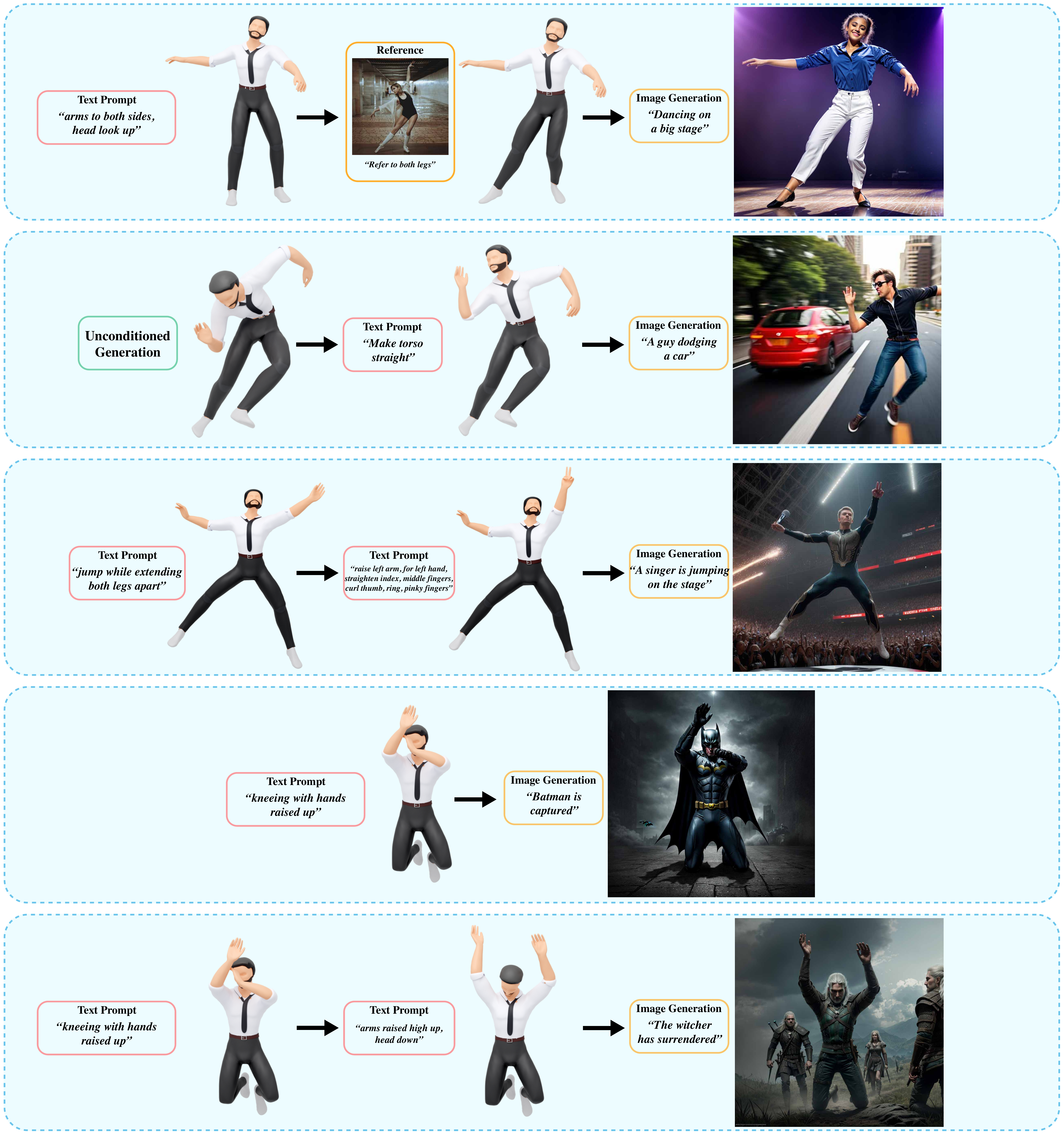}
    \caption{\textbf{Experimental Workflow:}
    More results for our experimental workflow for easy pose generation and modification, as well as its downstream image generation results.
    }
    \label{fig:supp_workflow}
\end{figure*}

\begin{figure*}
    \centering
    \includegraphics[width=0.85\linewidth]{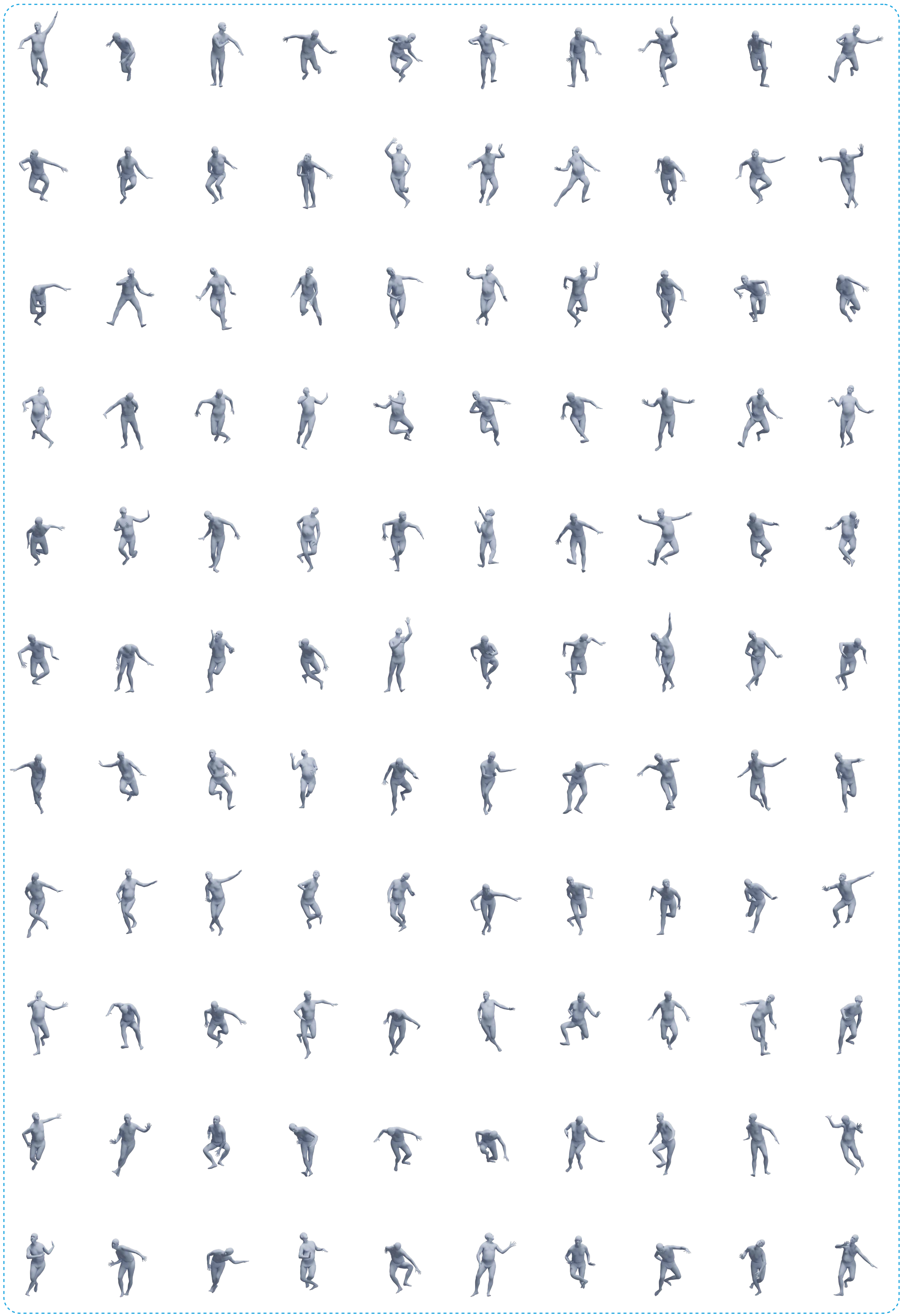}
    \caption{\textbf{Random Sampling:}
    Additional results from sampling within QPoser's latent space further demonstrate the viability and diversity of the generated poses, underscoring its expressiveness and correctness.
    }
    \label{fig:supp_uncon}
\end{figure*}

We showcase additional examples of our experimental workflow in Figure \ref{fig:supp_workflow}, emphasizing the practicality of having an expressive and controllable pose prior. Furthermore, we demonstrate the expressiveness and correctness of our pose prior by generating over 100 poses randomly using the method discussed in \ref{sec:supp_latent_sampling}, as depicted in Figure \ref{fig:supp_uncon}.

\end{document}